\documentclass[10pt, conference]{IEEEtran}

\usepackage{cite}
\usepackage{amsmath,amssymb,amsfonts}
\usepackage{algorithmic}
\usepackage{graphicx}
\usepackage{textcomp}
\usepackage{xcolor}
\usepackage{url}

\usepackage{subfig}
\newcommand{\DejaSansFont}{\usefont{T1}{DejaVuSans-TLF}{m}{n}}
\usepackage{overpic}
\usepackage{subfig}
\usepackage{booktabs}
\usepackage{flafter}
\def\BibTeX{{\rm B\kern-.05em{\sc i\kern-.025em b}\kern-.08em
    T\kern-.1667em\lower.7ex\hbox{E}\kern-.125emX}}

\begin{document}
\bstctlcite{IEEEexample:BSTcontrol}
\title{Reciprocity-Aware Convolutional Neural Networks for Map-Based Path Loss Prediction}

\author{
    \IEEEauthorblockN{Ryan G. Dempsey\IEEEauthorrefmark{1}\IEEEauthorrefmark{2}, Jonathan Ethier\IEEEauthorrefmark{1}\IEEEauthorrefmark{2}, Halim Yanikomeroglu\IEEEauthorrefmark{2}}\\
    \IEEEauthorblockA{\IEEEauthorrefmark{1}Communications Research Centre (CRC), Ottawa, ON, Canada:\\\{ryan.dempsey, jonathan.ethier\}@ised-isde.gc.ca}\\
    \IEEEauthorblockA{\IEEEauthorrefmark{2}Carleton University, Ottawa, ON, Canada:\\ryandempsey@cmail.carleton.ca, ethierjonathan@cunet.carleton.ca, halim@sce.carleton.ca}
}

\maketitle

\begin{abstract}
    Path loss modeling is a widely used technique for estimating point-to-point losses along a communications link from transmitter (Tx) to receiver (Rx). Accurate path loss predictions can optimize use of the radio frequency spectrum and minimize unwanted interference. Modern path loss modeling often leverages data-driven approaches, using machine learning to train models on drive test measurement datasets. Drive tests primarily represent downlink scenarios, where the Tx is located on a building and the Rx is located on a moving vehicle. Consequently, trained models are frequently reserved for downlink coverage estimation, lacking representation of uplink scenarios. In this paper, we demonstrate that data augmentation can be used to train a path loss model that is generalized to uplink, downlink, and backhaul scenarios, training using only downlink drive test measurements. By adding a small number of synthetic samples representing uplink scenarios to the training set, root mean squared error is reduced by \textgreater~8~dB on uplink examples in the test set.
\end{abstract}

\begin{IEEEkeywords}
Data augmentation, drive test measurements, path loss modeling, reciprocity.
\end{IEEEkeywords}

\section{Introduction}
\label{sec:intro}
Modern communications networks rely on path loss predictions to facilitate efficient use of the radio frequency spectrum while avoiding harmful interference. By providing a generic metric of losses along a point-to-point~(P2P) communications link from transmitter~(Tx) to receiver~(Rx), path loss provides a flexible quantity that is a stepping stone to many quality-of-service metrics. Path loss estimation is crucial for both uplink and downlink scenarios, particularly in the context of interference analysis. Path loss along a given communications link is primarily dependent on distance, frequency, and obstructions around the path. This can include blockages to the line of sight~(LOS), as well as obstructions within the first Fresnel zone~\cite{balanis}. A wireless signal's interactions with obstructions in the first Fresnel zone can affect path loss in intricate ways. Therefore, accounting for these obstructions in path loss modeling is paramount.

Modern path loss models rely on empirical, data-driven approaches to achieve low root mean squared error (RMSE) during inference in unseen geographic regions. With enough data, neural networks can be efficient for modeling complex, nonlinear relationships~\cite{universal-fcn, deeplearning-ref}.  While data-driven approaches can be helpful for such relationships, the resulting models can only be as powerful as their training data. More diverse training data can often help neural networks reduce RMSE in both training and inference~\cite{deeplearning-ref}.

\begin{figure}
    \centering
    \includegraphics[width=\linewidth]{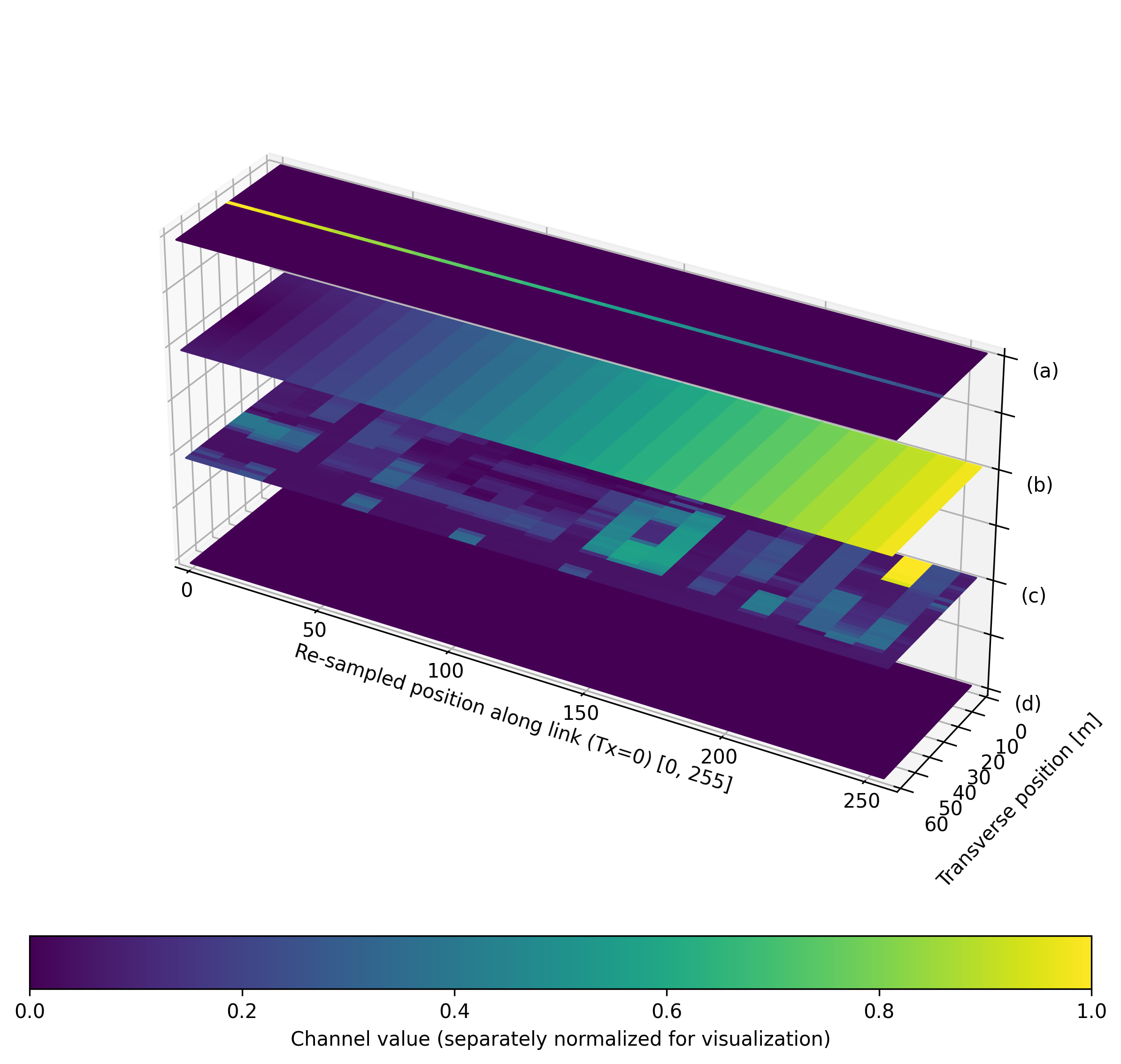}
    \caption[All input channels (re-sampled and normalized) visualized for a 411~m link]{All input channels (re-sampled and normalized) visualized for a 411~m link. (a) Direct path. (b) Distance from Tx to each pixel. (c) Surface. (d)~Frequency. Channels are separately normalized for visualization.}
    \label{fig:path-profile-normed-stacked}
\end{figure}

Our previous work~\cite{awpl-cite} introduced a convolutional neural network (CNN)-based method to automatically extract high-dimensional spatial features from 2-D obstruction height maps. These obstruction height maps were extracted from open-source digital surface model (DSM) data~\cite{uk-dsm-dtm}, and formed one of four feature channels~\cite{deeplearning-ref} input to the CNN. The other three channels were: the height of the 3-D direct path from Tx to Rx, a map of all 2-D distances from the Tx, and the frequency. The CNNs rely solely on channel-based inputs, since representing frequency and distance as scalar inputs yields inconsistent test RMSE~\cite{aps-cite}. The 2-D distance channel is used to encode locality for the CNN, similarly to~\cite{coordconv}. An example of a normalized input to the CNN can be seen in Fig.~\ref{fig:path-profile-normed-stacked}.  As discussed in~\cite{awpl-cite}, the direct path channel is computed by adding terrain height to the two antenna heights, and interpolating the line between the resulting values. Normalization details for each channel can be found in~\cite{awpl-cite}. Using a rigorous six-city cross-validation, we found automatic feature extraction using CNNs to be effective, yielding a mean RMSE of 7.35~dB across all folds.

The extensive cross-validations conducted in~\cite{awpl-cite} and~\cite{aps-cite} shows that the 2-D CNN provides a highly accurate model for coverage estimation. When the locations of the Tx and Rx are known, it can provide path loss predictions with 5-8~dB RMSE in geographically independent holdouts. However, it is unclear whether the CNN provides an accurate model of the underlying physics. Since CNN training is a data-driven, empirical approach, there is no guarantee that the CNN has implicitly learned any electromagnetic principles. Since each Tx and Rx are presented to the CNN in consistent locations, and the Tx is often more elevated than the Rx in the training data, it is worth studying whether the CNN has implicitly learned the reciprocal properties of path loss. Reciprocity in this context means that path loss should be identical regardless of which antenna is Tx and which antenna is Rx, assuming all environmental obstructions are passive and non-magnetic~\cite{balanis}.

Knowing whether the CNN is implicitly aware of reciprocity is an important distinction for operating the model in practice. If the CNN is unaware, the model may be considered primarily for use as a \textit{coverage model}, which is often used to generate path loss for several Rx points given a single Tx. In this case, the Tx would be a base station (BS), and the Rx would be user equipment (UE), as is often true in drive test data. Service providers and regulators could be confident in the path loss predictions based on the test RMSE, since these scenarios are similar to the training and test sets (BS-to-UE). It would not be recommended for use in the other three scenarios (BS-to-BS, UE-to-UE, or UE-to-BS). Conversely, if the CNN is aware of reciprocity, it can be considered more suitable for a wider range of scenarios, since its RMSE was low in such cases.

This paper investigates the use of data augmentation for training map-based path loss models that are aware of reciprocity, providing low RMSE for multiple link scenarios. Unlike~\cite{awpl-cite} and~\cite{aps-cite} which are limited to downlink scenarios, this paper evaluates models in more diverse links. Section~\ref{sec:data} describes two measurement datasets used to train and evaluate models in various link scenarios. Section~\ref{sec:link-types} begins with an overview of the transformations applied to represent various scenarios, followed by testing on multiple scenarios not seen in training. Section~\ref{sec:training} describes the use of these transformations to augment the training, and is followed by Section~\ref{sec:discussion}, which discusses the results. Section~\ref{sec:conc} concludes the paper.

\section{Measurement Data}
\label{sec:data}
The CNNs were trained on open-source drive test measurements~\cite{open-uk} from the United Kingdom Office of Communications (UK Ofcom)~\cite{uk-ofcom}. The measurements comprise the following frequency bands: (449, 915, 1802, 2695, 3602, and 5850 MHz). In~\cite{awpl-cite} and~\cite{aps-cite}, we focused our studies on the following six drive tests for their availability of DSM data: Boston, London, Merthyr Tydfil, Nottingham, Southampton, and Stevenage. For our cross-validation, after all data preprocessing, we used a random 10\,000 measurements in each band from each drive test (for a total of 360\,000 measurements). Each cross-validation fold is conducted 10 times to assess consistency, using a random 2000 measurements in each band from each drive test, from the 10\,000 subset (for a total of 72\,000 measurements across training, validation, and test). Each fold holds out a single drive test, with the remaining five drive tests split between training and validation.

Holding out entire drive tests is a form of geographic cross-validation~\cite{gis-split}, providing a proper assessment of model generalization to unseen regions. Geographic cross-validation is necessary for this data since different regions can have varied statistics in their DSM data, and an individual region's DSM data can often identifiable characteristics that are common throughout the region. Therefore, training and testing within the same region(s) can provide an optimistic assessment of performance due to spatial autocorrelation~\cite{spatial-autocorrelation}. Holding out the drive tests ensures rigorous assessment by preventing data leakage between adjacent samples from training to test within a region. More information on data cleaning, cross-validation approach, and hyperparameters can be found in~\cite{awpl-cite}.

In this paper, we also use 3455~MHz measurements taken by Communications Research Centre Canada (CRC) across 15 measurement sites in Ottawa~\cite{crc-measurements} to evaluate models on BS-to-BS scenarios. Tx and Rx heights in this dataset are both 11~m above ground. The dataset comprises 71 measurements after completing the filtering steps described in~\cite{awpl-cite}. Table~\ref{tab:data-summary} summarizes both measurement datasets.

\begin{table*}[htb]
    \centering
    \caption{Summary of Measurement Data}
    \label{tab:data-summary}
    \begin{tabular}{lcc}
    \toprule
            & \textbf{Ofcom} & \textbf{CRC} \\
        \midrule
        Frequencies [MHz] & [449, 915, 1802, 2659, 3602, 5850]  & [3455] \\
        Signal type & Continuous wave & Frequency-modulated continuous wave \\
        Bandwidth [MHz] & - & 8 \\
        Regions & Six\textsuperscript{*} regions across the UK & Ottawa, Canada \\
        Measurement filtering & Above system noise floor by 6~dB margin\textsuperscript{**} & Below maximum measurable path loss \\
        Measurements used & 360\,000 & 71 \\
        Tx height [m] & 17\textsuperscript{\textdagger} & 11 \\
        Rx height [m] & 1.5 & 11 \\
        Scenario & BS-to-UE & BS-to-BS \\
        Example & Downlink & Backhaul \\
    \bottomrule
    \end{tabular}
    \\
    \vspace{0.1cm}
    \textsuperscript{*}Measurement data is also available for Scar Hill, but high-resolution DSM data was not available at the time of publication.\\
    \vspace{0.1cm}
    \textsuperscript{**}As per guidelines to reduce systematic measurement errors to 1~dB or lower~\cite{ofcom-data-prep}.\\
    \vspace{0.1cm}
    \textsuperscript{\textdagger}For all drive tests except London, which used a 25~m Tx.
\end{table*}

\section{Data Representations}
\label{sec:link-types}
The Ofcom data provides BS-to-UE measurements. These measurements will be used to represent multiple link scenarios in both training and evaluation. Conversely, the CRC data provides BS-to-BS measurements, to be used for evaluation.

\subsection{Ofcom Data Transformations}
\label{sec:transforms}
The Ofcom data provides BS-to-UE measurements. In this paper, we transform the data to represent the opposite scenario: UE-to-BS. To this end, we define the following two transformations:
\begin{itemize}
    \item Identity: These are the original path profiles. In the Ofcom dataset, this may be thought of as \textit{downlink} scenarios, where the Tx is fixed on a tower and the Rx is on a moving vehicle.
    \item Reflection: This reflects path profiles across both horizontal axes, resulting in a reversal of the Tx and Rx, with transverse orientation of the original geographical environment preserved. In the Ofcom dataset, this may be thought of as \textit{uplink} scenarios, by swapping the Tx and Rx of the original links (equivalent to a 180\textdegree\ rotation).
\end{itemize}

Reflection should provide identical path loss~\cite{balanis}, assuming the constituent materials of all obstructions are passive and non-magnetic. When considering any polarization, any changes are consistently mirrored by directly swapping the Tx and Rx in an identical environment. The relative orientation between the polarization and the reflecting obstructions remains consistent, and the cumulative effects of the obstructions do not change, resulting in identical path loss. This phenomenon is known as reciprocity.

\begin{figure}[htb]
    \centering
    \subfloat[]{
        \begin{minipage}{0.97\linewidth}
        \centering
            \begin{overpic}[width=0.97\linewidth, grid=false, trim=-0.5cm 1.75cm 0cm 0cm, clip]{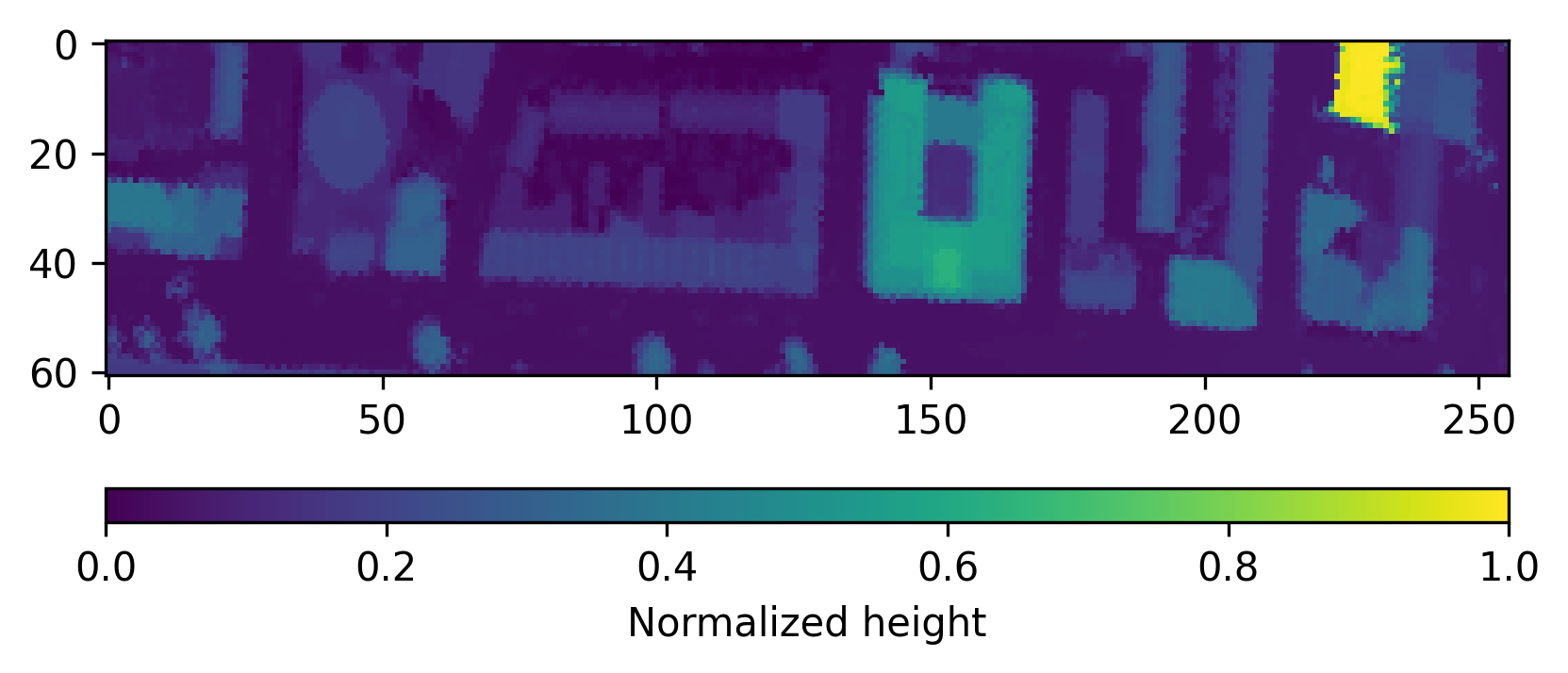}
                \put(1, -7){\rotatebox{90}{\DejaSansFont \scriptsize Transverse position [m]}}
                \put(19, -1){{\DejaSansFont \scriptsize Re-sampled position along link (Tx=0) [0, 255]}}
            \end{overpic}
        \hfill
        \\
        \vspace{0.5cm}
            \begin{overpic}[width=0.98\linewidth, grid=false, trim=-0.5cm 0cm 0cm 4cm, clip]{figs/London_449_14139_subsampled_normalized.png}
            \end{overpic}
        \end{minipage}
    }
    \hfil
    \subfloat[]{
        \begin{minipage}{0.97\linewidth}
        \centering
            \begin{overpic}[width=0.97\linewidth, grid=false, trim=-0.5cm 1.75cm 0cm 0cm, clip]{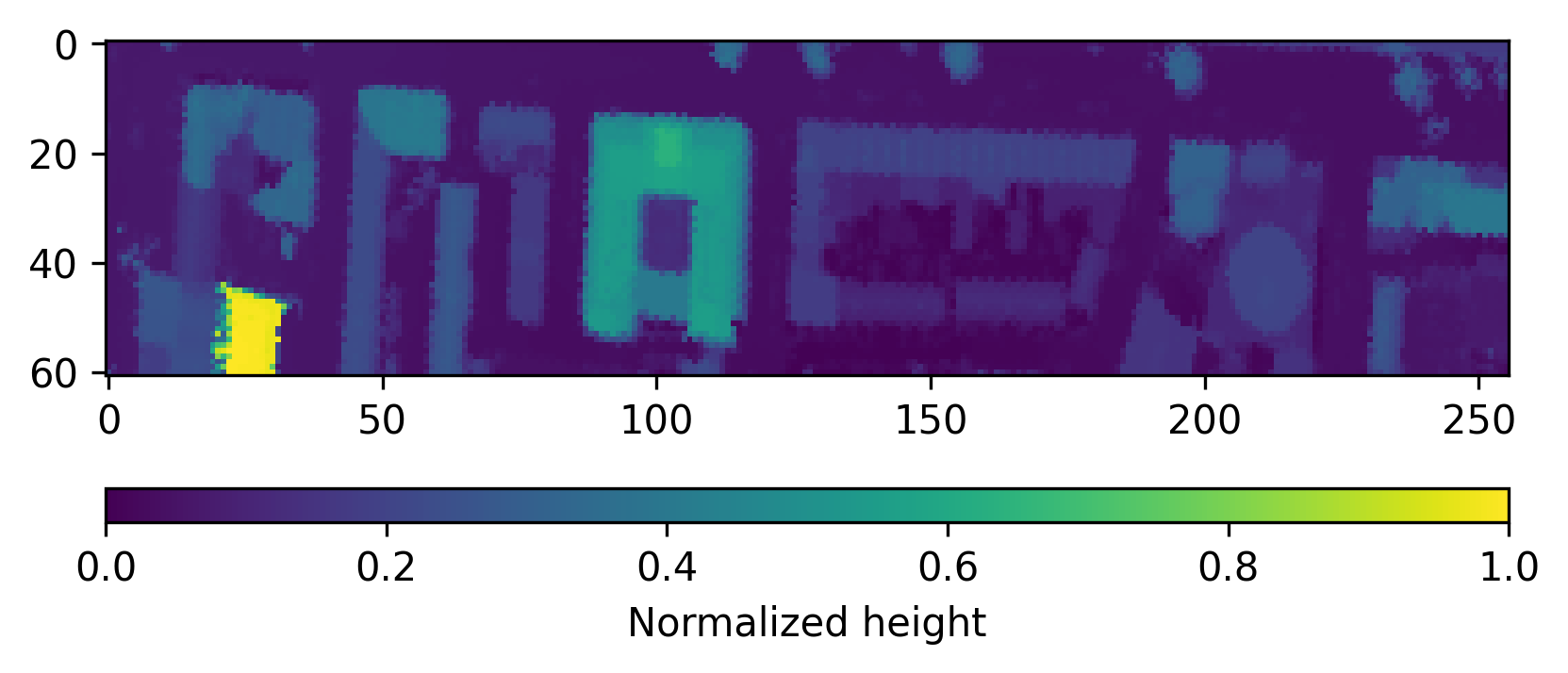}
                \put(1, -7){\rotatebox{90}{\DejaSansFont \scriptsize Transverse position [m]}}
                \put(19, -1){{\DejaSansFont \scriptsize Re-sampled position along link (Tx=0) [0, 255]}}
            \end{overpic}
        \hfill
        \\
        \vspace{0.5cm}
            \begin{overpic}[width=0.97\linewidth, grid=false, trim=-0.5cm 0cm 0cm 4cm, clip]{figs/London_449_14139_subsampled_normalized_fullreflect.png}
            \end{overpic}
        \end{minipage}
    }
    \caption[Path profile transformations visualized in 2-D using re-sampled and normalized CNN inputs]{Path profile transformations visualized in 2-D using re-sampled and normalized CNN inputs. (a) Identity (original path profile). (b) Reflection.}
    \label{fig:reflections}
\end{figure}

\begin{figure}
    \centering
    \subfloat[]{
    \includegraphics[width=0.9\linewidth]{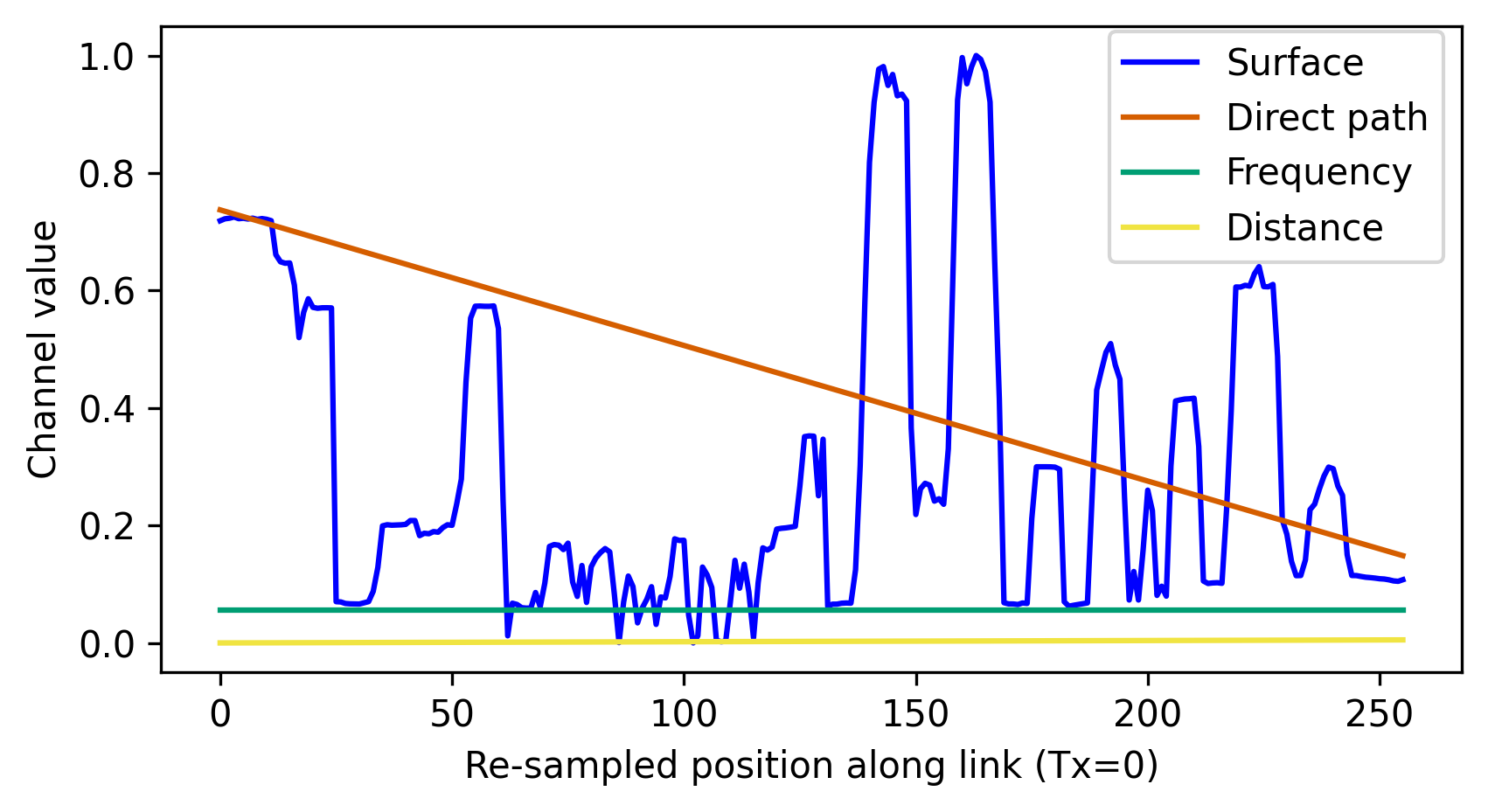}}
    \hfil
    \subfloat[]{\includegraphics[width=0.9\linewidth]{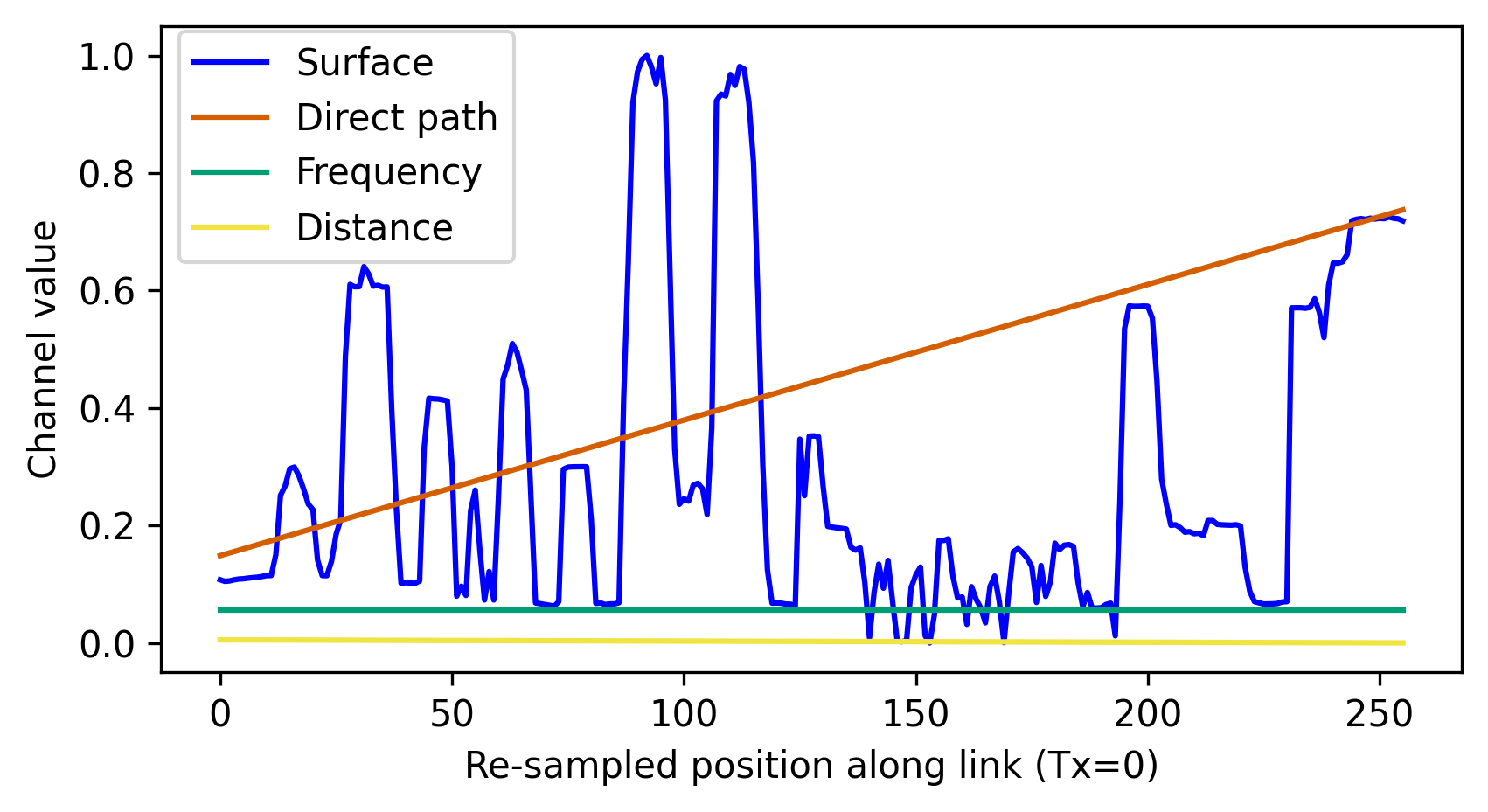}  
    }
    \caption[Reflection visualized in 1-D using re-sampled and normalized CNN inputs]{Reflection visualized in 1-D using re-sampled and normalized CNN inputs. (a) Identity (original path profile). (b) Reflected path profile.}
    \label{fig:reflections-1d}
\end{figure}

The transformation is visualized in 2-D in Fig.~\ref{fig:reflections} and in 1-D in Fig.~\ref{fig:reflections-1d}. These figures show the re-sampled and normalized inputs to the CNN. In this case, the Tx and Rx are oriented in the center of the path profile, on the left (position 0) and right (position 255) respectively, in the original array representation.

\subsection{CRC Measurement Data}
\begin{figure}
    \centering
    \subfloat[]{
        \begin{minipage}{0.97\linewidth}
        \centering
            \begin{overpic}[width=0.97\linewidth, grid=false, trim=-0.5cm 1.75cm 0cm 0cm, clip]{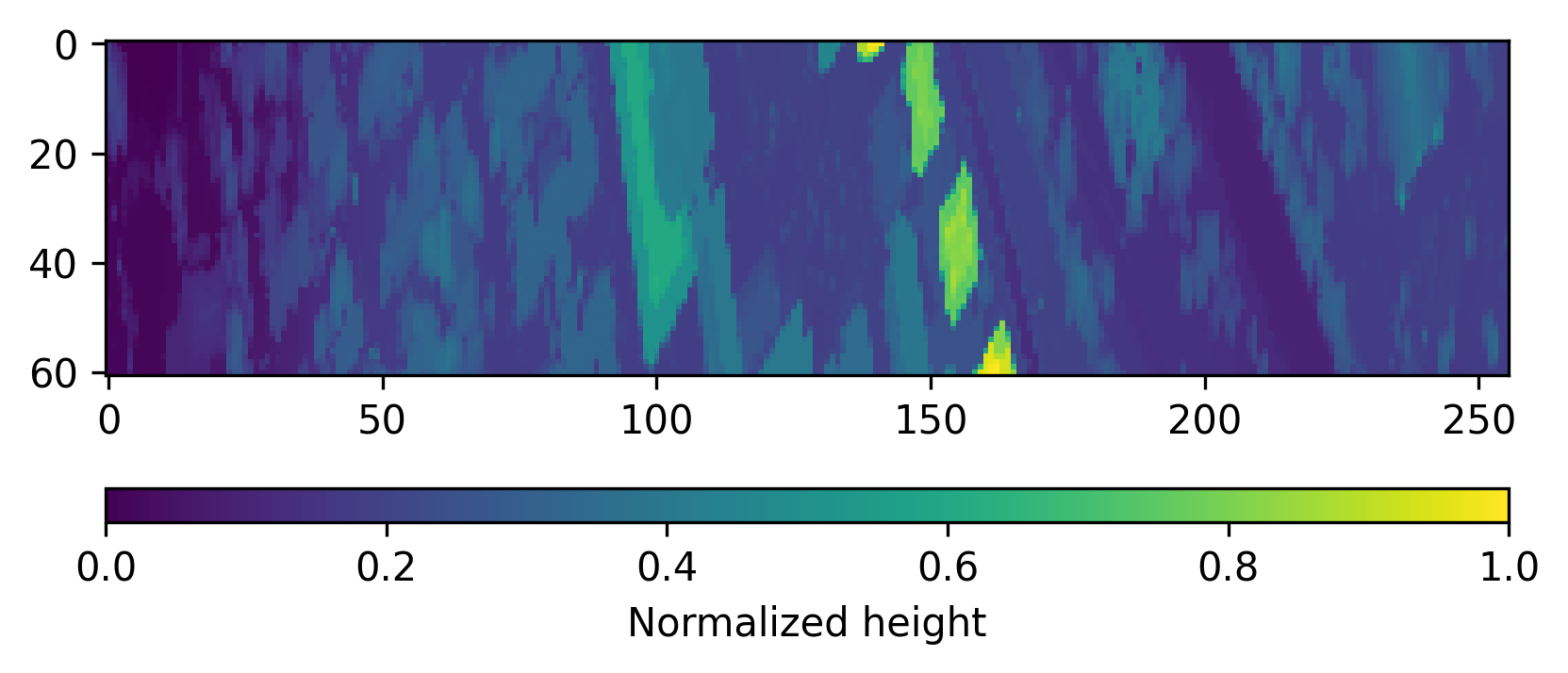}
                \put(1, -7){\rotatebox{90}{\DejaSansFont \scriptsize Transverse position [m]}}
                \put(19, -1){{\DejaSansFont \scriptsize Re-sampled position along link (Tx=0) [0, 255]}}
            \end{overpic}
        \hfill
        \\
        \vspace{0.5cm}
            \begin{overpic}[width=0.98\linewidth, grid=false, trim=-0.5cm 0cm 0cm 4cm, clip]{figs/Ottawa_3455_85.png}
            \end{overpic}
        \end{minipage}
    }
    \hfil
    \subfloat[]{
        \includegraphics[width=0.9\linewidth]{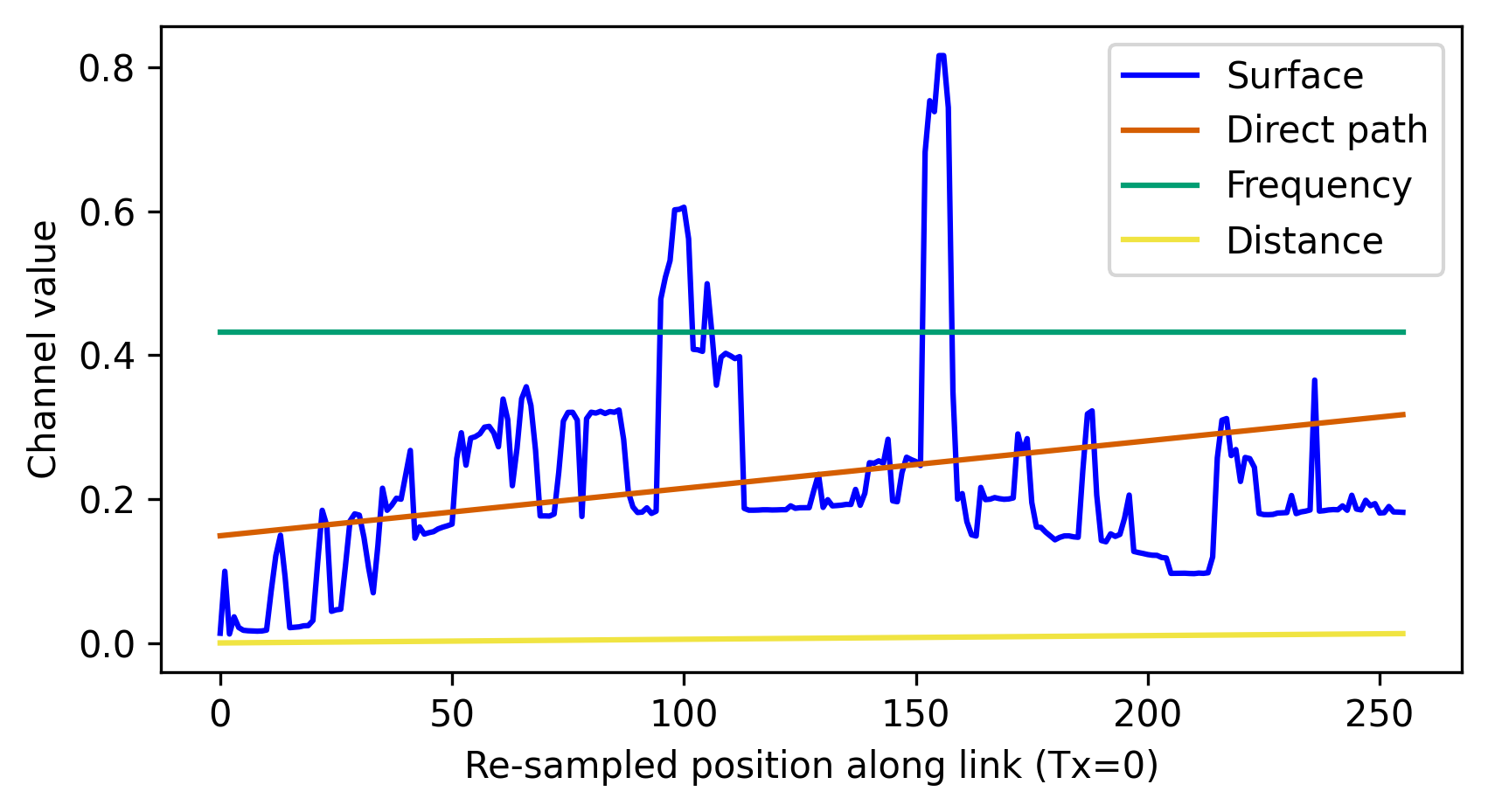}
    }
    \caption[Ottawa path profile visualized using re-sampled and normalized CNN inputs]{Ottawa path profile visualized using re-sampled and normalized CNN inputs. (a) Surface visualized in 2-D. (b) All channels visualized in 1-D.}
    \label{fig:crc-path-profile}
\end{figure}

The CRC dataset provides measurements of BS-to-BS scenarios, where both Tx and Rx are antennas fixed at a tall height. Fig.~\ref{fig:crc-path-profile} shows a re-sampled and normalized path profile derived from the CRC data, for a 1~km link. Table~\ref{tab:path-profile-scenarios} summarizes the dataset representations for all path profiles.

\begin{table}
    \centering
    \caption{Datasets Used to Represent Various Link Scenarios}
    \label{tab:path-profile-scenarios}
    \begin{tabular}{cccc}
    \toprule
        \textbf{Scenario} & \textbf{Example} & \textbf{Data Source} & \textbf{Transformation} \\
    \midrule
        BS-to-UE & Downlink & Ofcom & Identity \\
        UE-to-BS & Uplink & Ofcom & Reflection \\
        BS-to-BS & Backhaul & CRC & N/A \\
        UE-to-UE & Internet of things & \multicolumn{2}{c}{Out of scope} \\
    \bottomrule
    \end{tabular}
\end{table}

\subsection{Model Evaluation on Multiple Test Scenarios}
Reflection is conducted on the test sets (holdouts) from the original cross-validation from~\cite{awpl-cite}. The corresponding trained CNN is evaluated on each of these new transformed test sets (UE-to-BS), and compared with the original results (``identity'' transformation, BS-to-UE), via mean and standard deviation (SD) in Table~\ref{tab:lr-flipped-path-profiles}. Each CNN is also evaluated on the BS-to-BS samples from Ottawa~\cite{crc-measurements}.

\begin{table}[!tp]
\centering
\caption{Original Model Test RMSE [\textnormal{d}B] on Various Test Sets}
\label{tab:lr-flipped-path-profiles}
\begin{tabular}{lcccc|cc}
\toprule
        & \multicolumn{2}{c}{\textbf{BS-to-UE}} & \multicolumn{2}{c}{\textbf{UE-to-BS}} & \multicolumn{2}{|c}{\textbf{Ottawa BS-to-BS}}\\
       \textbf{Holdout} & Mean & SD & Mean & SD & Mean & SD\\
\midrule
Boston          & 7.18 & 0.27 & 14.39 & 4.51 & 7.00 & 0.44 \\
London          & 7.75 & 0.47 & 14.30 & 2.40 & 7.48 & 0.50 \\
Merthyr Tydfil  & 8.07 & 0.39 & 14.08 & 2.67 & 7.17 & 0.21 \\
Nottingham      & 6.99 & 0.23 & 16.88 & 4.65 & 7.62 & 0.36 \\
Southampton     & 6.35 & 0.42 & 17.06 & 3.14 & 7.52 & 0.47 \\
Stevenage       & 7.73 & 0.21 & 20.49 & 3.80 & 7.18 & 0.30 \\
\midrule
Mean            & 7.35 & 0.33 & 16.20 & 3.53 & 7.33 & 0.38 \\
\bottomrule
\end{tabular}
\end{table}

The high RMSEs of the reflected holdouts indicate that the model is tuned for the specific orientation between BS and UE that was present in the training set (downlink). In the training set, the majority of samples contained a tall Tx on one side of the image and a short Rx on the other. Though the exact slope of the direct path varies due to terrain, the general trend is that the Tx is much higher than the Rx. This is a result of both the Ofcom dataset (drive tests primarily use a short Rx and a tall Tx) as well as the training methodology (Tx and Rx in consistent orientation). The result is a training set with an over-representation of tall, BS-style antennas on one consistent side and shorter antennas consistently on the other, and under-representation of the reverse.

Interestingly, the models perform reasonably well on the CRC data. Despite holding out different UK regions from training, RMSEs in Ottawa are generally lower than their UK counterparts. Despite the small sample size, this may be indicative of the fact that drive tests generally are a lot noisier than BS-to-BS links, with more dynamic scenarios and generally more complex obstructions.

\section{Training Set Augmentation}
\label{sec:training}
The data-driven nature of the proposed method has resulted in some physically inaccurate estimations. The trained models provide high RMSE only for uplink scenarios, where the Rx is a BS and the Tx is analogous to a UE (in this case a drive test antenna on a vehicle). This is the result of training primarily on downlink scenarios. However, the data-driven nature can also be used advantageously: By supplementing the training and validation sets with samples representing uplink scenarios (i.e. samples undergoing reflection), the augmented training set may implicitly teach the model the reciprocity of path loss.

\begin{figure}
    \centering
    \subfloat[]{
        \includegraphics[width=0.47\textwidth]{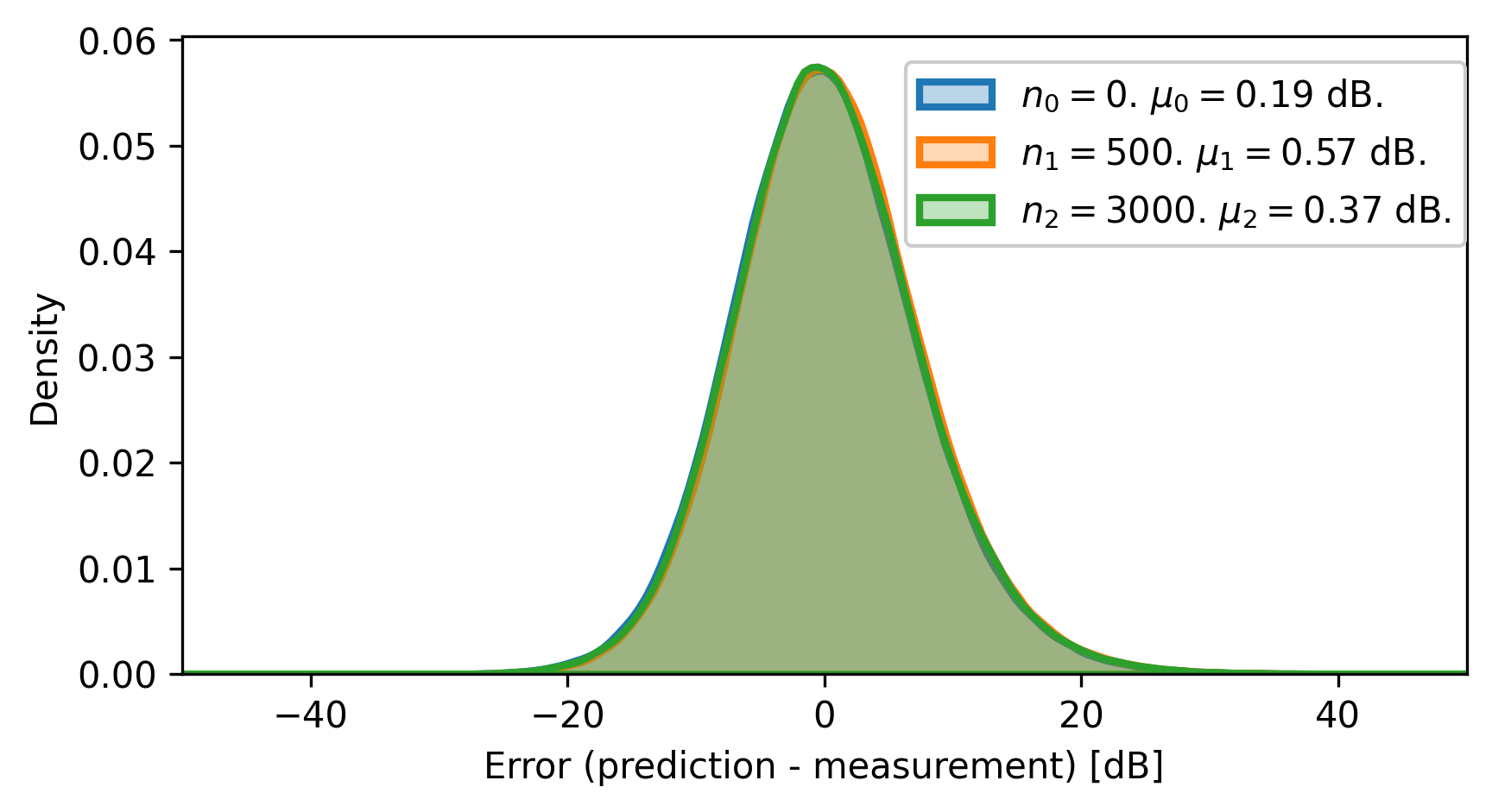}
    }
    \hfil
    \subfloat[]{
        \includegraphics[width=0.47\textwidth]{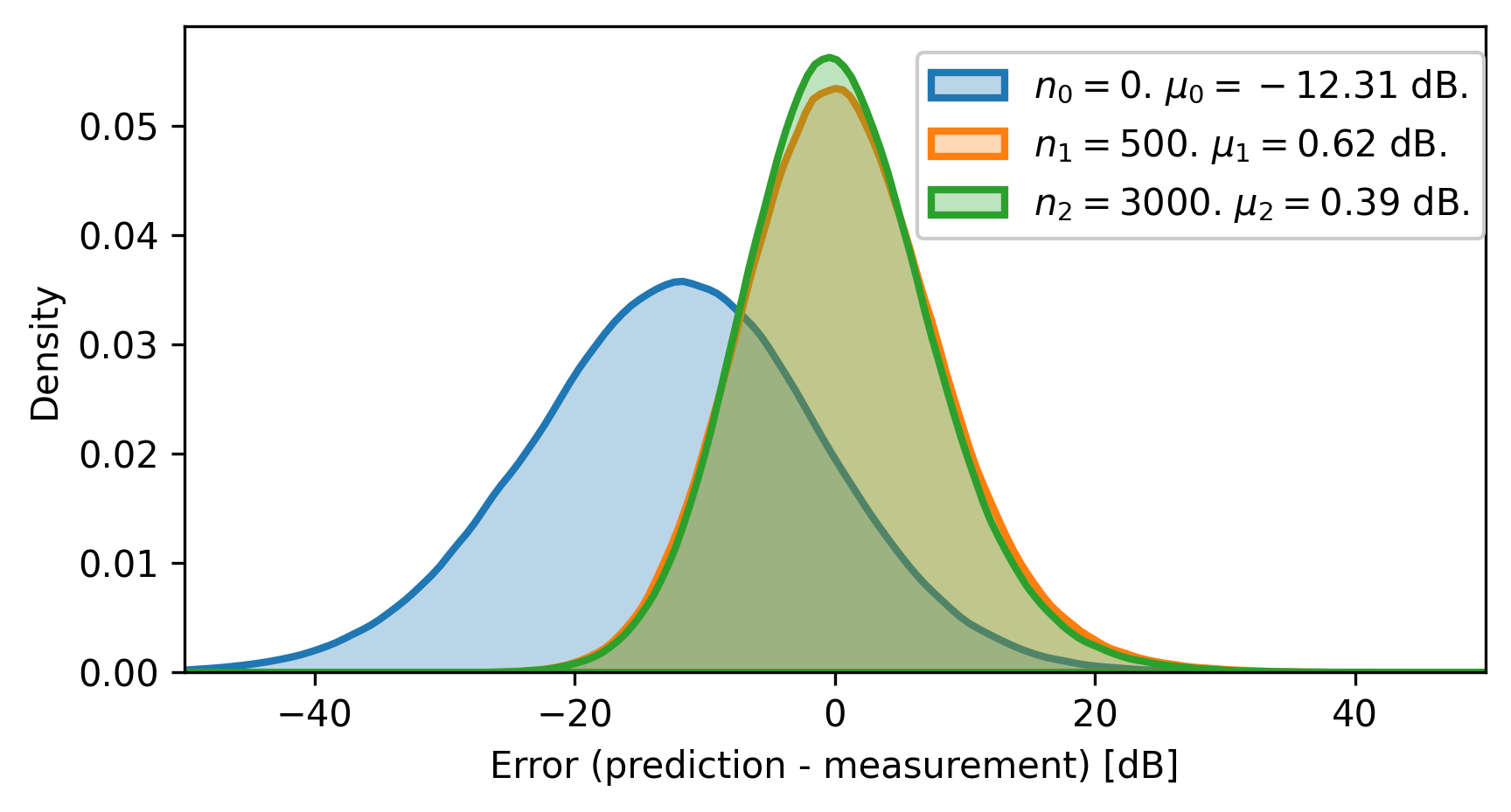}
        \label{fig:uplink-hist}
    }

    \hfil
    \subfloat[]{
        \includegraphics[width=0.47\textwidth]{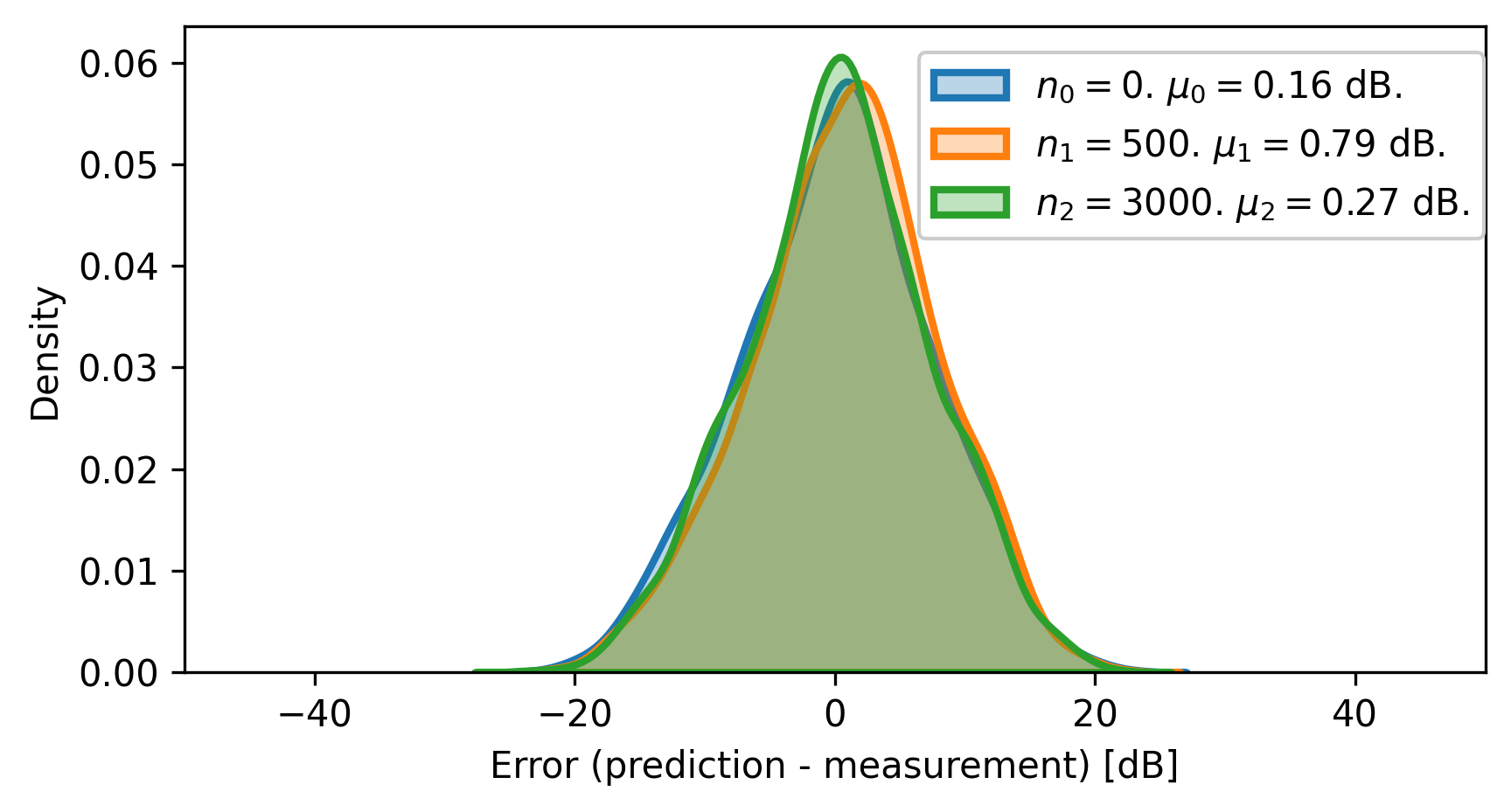}
    }
    \caption{KDE of prediction bias. (a) Identity (BS-to-UE). (b) Reflection (UE-to-BS). (c) CRC data (BS-to-BS).}
    \label{fig:hist}
\end{figure}

The model is re-trained with augmented training/validation sets. $n$ samples from each of the five training/validation cities are duplicated and reflected for multiple values of $n$ (resulting in 12\,000 + $n$ samples per city between training and validation). This is employed for varying values of $n$ (no augmentation, 4\%, and 25\%). As described in~\cite{awpl-cite}, each cross-validation is conducted 10 times.  The holdouts use 12\,000 samples from the held out drive test for both the reflected and original test sets. The resulting models are evaluated on both the original and reflected test sets, as well as the CRC data. The results are shown in Tables~\ref{tab:flipped-path-profiles},~\ref{tab:flipped-path-profiles-flipped-test}, and~\ref{tab:ottawa-test} respectively. The results are summarized in Table~\ref{tab:n-augmentations}. Finally, Fig.~\ref{fig:hist} visualizes the error distribution across all holdouts using kernel density estimation (KDE).

\begin{table}[!ht]
\centering
\caption{RMSE [\textnormal{d}B] on Identity Test Sets with Augmented Training}
\label{tab:flipped-path-profiles}
\begin{tabular}{lcccccccc}
\toprule
        & \multicolumn{2}{c}{$\boldsymbol{n_0=0}$} & \multicolumn{2}{c}{$\boldsymbol{n_1=500}$} & \multicolumn{2}{c}{$\boldsymbol{n_2=3000}$}\\
       \textbf{Holdout} & Mean & SD & Mean & SD & Mean & SD\\
\midrule
Boston          & 7.18 & 0.27 & \textbf{6.98} & 0.59 & 7.04 & 0.27\\
London          & \textbf{7.75} & 0.47 & 8.40 & 0.82 & 8.21 & 0.62\\
Merthyr Tydfil  & 8.07 & 0.39 & \textbf{7.75} & 0.29 & 8.21 & 0.50\\
Nottingham      & 6.99 & 0.23 & 6.86 & 0.18 & \textbf{6.78} & 0.20\\
Southampton     & 6.35 & 0.42 & 6.40 & 0.37 & \textbf{6.25} & 0.26\\
Stevenage       & 7.73 & 0.21 & \textbf{7.54} & 0.23 & 7.67 & 0.32\\
\midrule
Mean            & 7.35 & 0.33 & \textbf{7.32} & 0.41 & 7.36 & 0.36 \\
\bottomrule
\end{tabular}
\end{table}

\begin{table}[!htbp]
\centering
\caption{RMSE [\textnormal{d}B] on Reflected Test Sets with Augmented Training}
\label{tab:flipped-path-profiles-flipped-test}
\begin{tabular}{lcccccccc}
\toprule
        & \multicolumn{2}{c}{$\boldsymbol{n_0=0}$} & \multicolumn{2}{c}{$\boldsymbol{n_1=500}$} & \multicolumn{2}{c}{$\boldsymbol{n_2=3000}$}\\
       \textbf{Holdout} & Mean & SD & Mean & SD & Mean & SD\\
\midrule
Boston          & 14.39 & 4.51 & 7.28 & 0.36 & \textbf{6.95} & 0.23\\
London          & 14.30 & 2.40 & 8.80 & 0.73 & \textbf{8.34} & 0.48\\
Merthyr Tydfil  & 14.08 & 2.67 & 8.71 & 0.36 & \textbf{8.64} & 0.51\\
Nottingham      & 16.88 & 4.65 & 7.40 & 0.18 & \textbf{6.76} & 0.11\\
Southampton     & 17.06 & 3.14 & 6.67 & 0.28 & \textbf{6.31} & 0.28\\
Stevenage       & 20.49 & 3.80 & 7.68 & 0.33 & \textbf{7.53} & 0.27\\
\midrule
Mean            & 16.20 & 3.53 & 7.76 & 0.37 & \textbf{7.42} & 0.31\\
\bottomrule
\end{tabular}
\end{table}

\begin{table}[!htbp]
\centering
\caption{RMSE [\textnormal{d}B] on BS-to-BS Test Set with Augmented Training}
\label{tab:ottawa-test}
\begin{tabular}{lcccccccc}
\toprule
        & \multicolumn{2}{c}{$\boldsymbol{n_0=0}$} & \multicolumn{2}{c}{$\boldsymbol{n_1=500}$} & \multicolumn{2}{c}{$\boldsymbol{n_2=3000}$}\\
       \textbf{Original Holdout} & Mean & SD & Mean & SD & Mean & SD\\
\midrule
Boston          & 7.00 & 0.44 & 6.91 & 0.46 & \textbf{6.83} & 0.42 \\
London          & 7.48 & 0.50 & 7.47 & 0.45 & \textbf{7.34} & 0.36 \\
Merthyr Tydfil  & 7.17 & 0.21 & 7.10 & 0.38 & \textbf{6.73} & 0.31 \\
Nottingham      & 7.62 & 0.36 & 7.15 & 0.36 & \textbf{7.14} & 0.41 \\
Southampton     & 7.52 & 0.47 & \textbf{7.32} & 0.44 & 7.61 & 0.28 \\
Stevenage       & 7.18 & 0.30 & 6.99 & 0.41 & \textbf{6.90} & 0.33 \\
\midrule
Mean            & 7.33 & 0.38 & 7.16 & 0.42 & \textbf{7.09} & 0.35 \\
\bottomrule
\end{tabular}
\end{table}

\begin{table}[!htbp] 
\centering
    \caption{Mean Test RMSE [\textnormal{d}B] with Augmented Training Runs}
    \label{tab:n-augmentations}
    \begin{tabular}{rcccccc}
    \toprule
           & \multicolumn{2}{c}{$\boldsymbol{n_0=0}$} & \multicolumn{2}{c}{$\boldsymbol{n_1=500}$} & \multicolumn{2}{c}{$\boldsymbol{n_2=3000}$}\\
           \textbf{Test Set} & Mean & SD & Mean & SD & Mean & SD \\
    \midrule
         BS-to-UE & 7.35 & 0.33 & \textbf{7.32} & 0.41 & 7.36 & 0.36 \\
         UE-to-BS & 16.20 & 3.53 & 7.76 & 0.37 & \textbf{7.42} & 0.31 \\
         BS-to-BS & 7.33 & 0.38 & 7.16 & 0.42 & \textbf{7.09} & 0.35 \\    
    \bottomrule
    \end{tabular}
\end{table}

\section{Discussion}
\label{sec:discussion}
The data augmentation has had a significant effect on test RMSE. Adding just 500 simulated uplink samples to each city (just 4\% of the original training/validation set) reduces uplink RMSE from 16.20~dB to 7.76~dB. Even the mean RMSE on the unaltered (downlink) test set is reduced from 7.35~dB to to 7.32~dB, likely benefiting from the increased sample diversity. Mean RMSE on the CRC (BS-to-BS) test set is also improved with training augmentation, from 7.33~dB to 7.16~dB. Increasing the number of transformed training/validation samples to 3000 per city (25\% of the original data) incrementally improves the performance on the transformed test set, to 7.42~dB mean RMSE. While the RMSE on the unaltered holdouts stays roughly the same as $n_0$ (within 0.01~dB), RMSE on the BS-to-BS test set is improved to 7.09~dB.

Fig.~\ref{fig:hist} demonstrates that without training representation, path loss is underestimated in uplink scenarios. One potential explanation is that the majority of obstructions are near the Rx in the downlink scenarios. The Tx is often above the nearby buildings while the Rx is surrounded by buildings and trees. This type of scenario being over-represented in training may have caused the models to underemphasize the area around the Tx while focusing on the area around the Rx. Consequently, the model may have overlooked the obstructions near the Rx in the uplink samples, resulting in a much lower predicted path loss than reality. Augmenting 4\% of training samples appears to have resolved the issue by allowing the model to learn that obstructions near both Tx and Rx can impact path loss. The $n_1$ models have a mean error of 0.62~dB on the reflected test samples---a significant improvement over $-$12.31~dB.

The results show that a small increase in training diversity can have a profound impact on model generalization, not only to the types of samples added but also to other types of samples. However, the number of augmented samples may be optimized with a dedicated study to minimize both uplink and downlink RMSE. In any case, the results indicate that this model can easily be converted from a coverage model (where the Tx is a known BS) to a generic P2P path loss model that works well for downlink, uplink, and backhaul scenarios, by adding a very small number of uplink samples to training.

\begin{figure}[!htbp]
    \centering

        \includegraphics[width=0.48\textwidth]{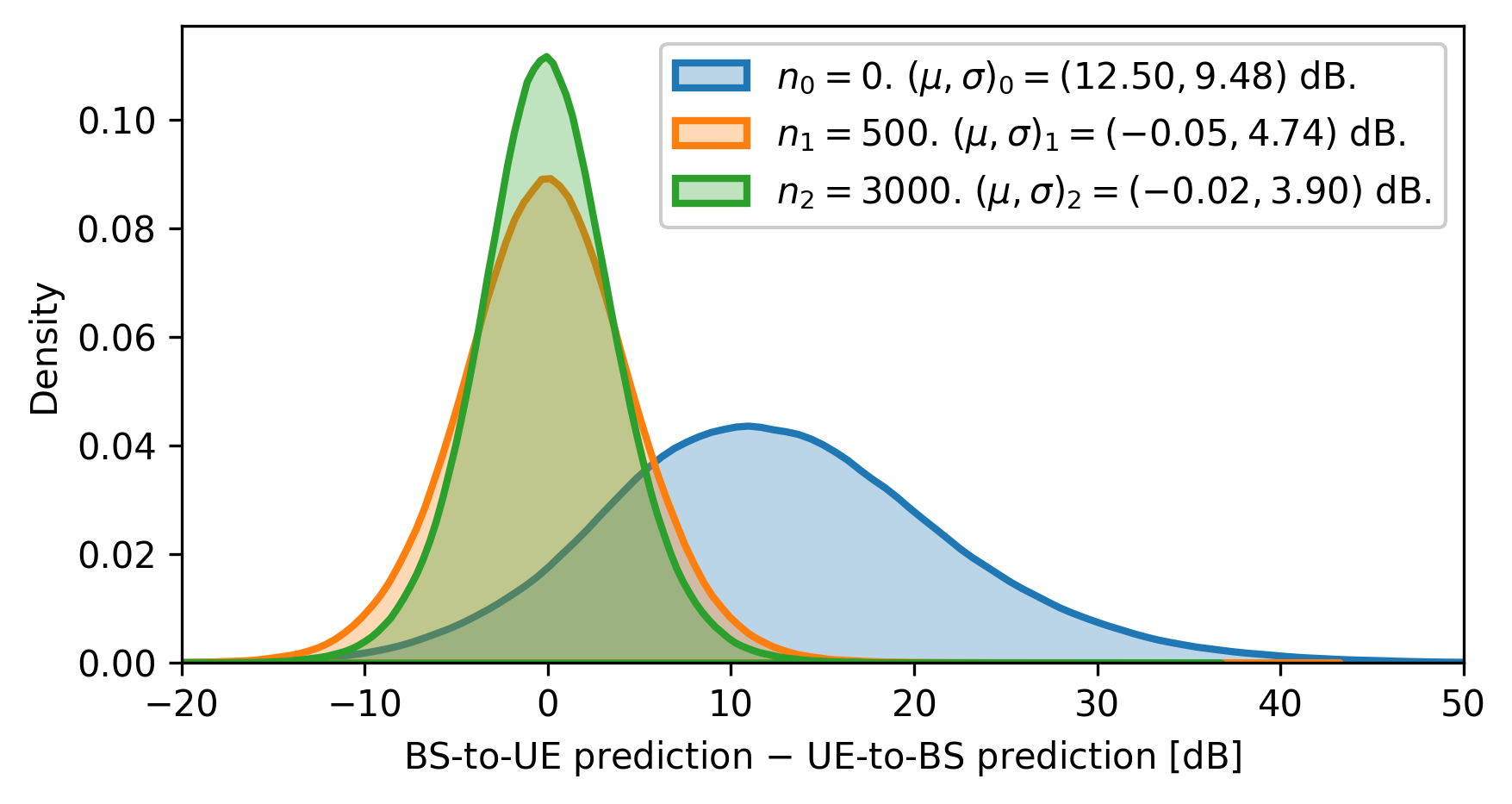}

    \caption[KDE of prediction on downlink minus prediction on uplink]{KDE of prediction on downlink minus prediction on uplink.}
    \label{fig:hist-recip}
\end{figure}

The results discussed thus far showcase the effect of training data augmentation on generalization, by comparing test RMSE. However, the original purpose of this study was to examine whether the CNNs can accurately model reciprocity. Therefore, it is beneficial to examine the individual predictions to see how well they align between downlink and uplink scenarios. Fig.~\ref{fig:hist-recip} shows the distribution of the difference between the identity prediction and the reflected prediction for each test link. For an ideal model with perfect reciprocity, the distribution would be a unit impulse at $x=0$, indicating that each sample has the same prediction in both directions.

The original models have a high mean and SD, showing both a lack of consistency and also a tendency to predict higher path loss in BS-to-UE samples than the reverse, which is consistent with the fact that path loss is under-predicted in UE-to-BS samples. The first training data augmentation decreases the mean difference from 12.50~dB to $-$0.05~dB, and the SD from 9.48~dB to 4.74~dB, reinforcing the power of training representation. The second training data augmentation has a more subtle yet still relevant improvement, to a mean and SD of $-$0.02~dB and 3.90~dB, respectively. The improvements provide diminishing returns, but still show an incremental improvement by tightening the distribution as individual uplink and downlink predictions converge.

Notably, one of the four scenarios discussed in Section~\ref{sec:intro} was not considered: UE-to-UE. This scenario would be useful for modeling a plethora of vertical communications networks, such as vehicle-to-vehicle (V2V) communications or even vehicle to everything (V2X). The work is this paper suggests that a small number of such measurements added to training may sufficiently improve test RMSE. A dedicated study would yield more precise insights.

\section{Conclusion}
\label{sec:conc}
This paper leveraged data augmentation to generalize path loss models to both uplink and downlink scenarios, starting with only downlink measurements. It was shown that by augmenting just 4\% of the training and validation data with synthetic samples representing uplink scenarios, test RMSE on unseen uplink samples was reduced from 16.20~dB to 7.76~dB on average, without sacrificing performance on the original downlink scenarios nor backhaul scenarios. We also showed that individual predictions' reciprocity was improved, resulting in fewer samples with large discrepancies between predicted path loss in each direction.

Future work should investigate the discussed techniques in a wider variety of scenarios, including vertical use cases, such as vehicular communications. Additionally, this approach should be validated in inference using uplink measurements. This would provide increased confidence on the model's ability to handle varied communications scenarios, which may not always be truly reciprocal in practice. A dedicated study may also determine the optimal number of augmented samples to balance training times with test RMSE on multiple link configurations. Finally, future research could also explore explicitly enforcing reciprocity constraints via alternative model architectures and/or loss functions, and compare the effectiveness against the data augmentation methods presented here.

\section*{Acknowledgment}
The authors would like to thank Mathieu Châteauvert for his advice on radio propagation and machine learning.

\bibliographystyle{IEEEtran}
\bibliography{references}
\vfill

\end{document}